\documentclass{article}

\PassOptionsToPackage{numbers, square, sort}{natbib}
\usepackage[final]{neurips_2021}

\usepackage[utf8]{inputenc} %
\usepackage[T1]{fontenc}    %
\usepackage{hyperref}       %
\usepackage{url}            %
\usepackage{booktabs}       %
\usepackage{amsfonts}       %
\usepackage{nicefrac}       %
\usepackage{microtype}      %
\usepackage{xcolor}         %
\usepackage{graphicx} %
\usepackage{epsfig} %
\usepackage{amsmath} %
\usepackage{amssymb}  %
\usepackage{color}
\usepackage{ulem}
\usepackage{subcaption}
\usepackage{algorithm}
\usepackage{algpseudocode}
\usepackage{url}
\usepackage{hyperref}

\newcommand{\methodname}{VSDR}

\title{Validate on Sim, Detect on Real -- \\
Model Selection for Domain Randomization}

\author{%
  Gal Leibovich\thanks{Equal contribution}\\
  Intel Labs\\
  \texttt{gal.leibovich@intel.com}
  \And
  Guy Jacob\footnotemark[1]\\
  Intel Labs\\
  \texttt{guy.jacob@intel.com}
  \And
  Shadi Endrawis\\
  Intel Labs\\
  \texttt{shadi.endrawis@intel.com}
  \And
  Gal Novik\\
  Intel Labs\\
  \texttt{gal.novik@intel.com}
  \And
  Aviv Tamar\\
  Technion -- Israel Institute of Technology\\
  \texttt{avivt@technion.ac.il}
}

\begin{document}

\maketitle

\begin{abstract}
A practical approach to learning robot skills, often termed sim2real, is to train control policies in simulation and then deploy them on a real robot. Popular techniques to improve the sim2real transfer build on domain randomization (DR) -- training the policy on a diverse set of randomly generated domains with the hope of better generalization to the real world. Due to the large number of hyper-parameters in both the policy learning and DR algorithms, one often ends up with a large number of trained policies, where choosing the best policy among them demands costly evaluation on the real robot. In this work we ask -- \textit{can we rank the policies without running them in the real world?}
Our main idea is that a predefined set of real world data can be used to evaluate all policies, using out-of-distribution detection (OOD) techniques. In a sense, this approach can be seen as a `unit test' to evaluate policies before any real world execution. However, we find that by itself, the OOD score can be inaccurate and very sensitive to the particular OOD method. Our main contribution is a simple-yet-effective policy score that combines OOD with an evaluation in simulation. We show that our score -- \methodname\ -- can significantly improve the accuracy of policy ranking without requiring additional real world data.
We evaluate the effectiveness of \methodname\ on sim2real transfer in a robotic grasping task with image inputs. We extensively evaluate different DR parameters and OOD methods, and show that \methodname\ improves policy selection across the board. More importantly, our method achieves significantly better ranking, and uses significantly less data compared to baselines.
Project website is available at \url{https://sites.google.com/view/vsdr/home}
\end{abstract}

\section{Introduction}\label{s:intro}

Reinforcement learning (RL) is a popular method for learning various robotic skills~\cite{kober2013reinforcement}. Training and evaluating RL policies on a real robot, however, can be expensive, unsafe, and time consuming, and many recent RL advances were achieved by training policies in simulation and then transferring them to the real environment~\cite{pmlr-v87-golemo18a, tan2018simtoreal, peng2018simtoreal, Hwangbo_2019}. Simulation allows straightforward parallelization during training, and modern physical simulators can quickly and safely simulate realistic physical interactions~\cite{openai2019learning, openai2019solving}.

Since simulation and reality often cannot be perfectly matched, policies trained in simulation do not necessarily perform well in the real world -- a.k.a.~the \textit{sim-to-real gap}~\cite{james20163d}. Several techniques have been proposed to alleviate the gap~\cite{james2019simtoreal}, among which \textit{Domain Randomization (DR)}~\cite{tobin2017domain, peng2018simtoreal, sadeghi2017cad2rl} is particularly popular. In DR, instead of training on a single simulation environment, policies are trained on multiple domains with random variations in environment attributes such as textures, lighting, and dynamics. The hope is that policies trained using DR will be invariant to these variations, therefore mitigating the sim2real gap.

In practice, RL algorithms require tuning a large number of hyper-parameters~\cite{henderson2018deep}, including the learning rate, batch-size, discount factor, and network update frequency. DR adds additional hyper-parameters, including the elements in the environment that are randomized, the type of randomization, and the frequency in which randomization occurs. Common randomizations affect visual elements such as textures and lighting, or physical elements such as mass and friction. In this work we focus on DR hyper-parameters only, and specifically on visual randomizations, which are important for training vision-based policies for tasks such as grasping or placing.

Selecting the best hyper-parameters with respect to performance in simulation is easy, and can even be done automatically using cross validation techniques~\cite{zahavy2020self}. However, how can we select the best parameters with respect to \textit{real world} performance? This is the question we explore in this work. Formally, we seek to rank a set of policies based on their real world performance.

A straightforward solution is to run each policy in the real world enough times to obtain an accurate performance evaluation. When the number of policies is large, this approach can be very time consuming. For example, the manipulation task in our experiments requires 60-75 minutes of robot interaction to evaluate a single policy, and it is not uncommon to have hundreds of trained policies when using deep RL algorithms with DR.

\begin{figure}
    \centering
    \includegraphics[width=1.0\linewidth]{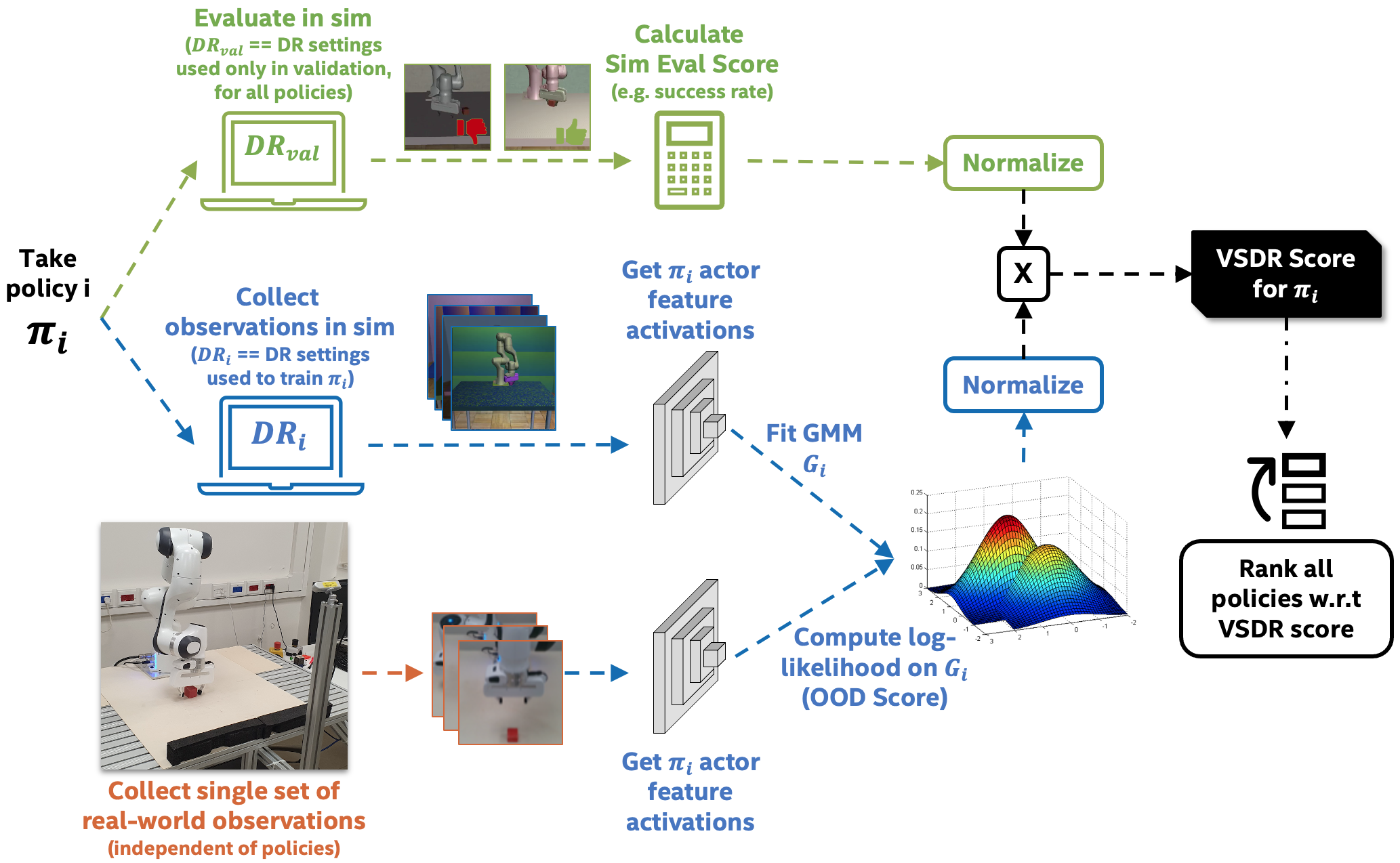}
    \caption{ We present an approach for selecting the best policies to deploy in the real-world, by combining prior evaluation in simulation with out-of-distribution detection in the real-world. Our method outperforms the compared baselines while requiring merely dozens of real-world observations. The bottom-left image in the diagram shows the lab setup used for real-world experiments, consisting of a Franka Panda robotic arm and an Intel RealSense D415 camera.}
    \label{fig:method_diagram}
\end{figure}

We posit that \textit{a fast cycle of policy evaluation}, which does not require running the policies on the real robot, is critical for practical applications of sim2real methods. Specifically our desiderata from the evaluation method are:
\begin{enumerate}
    \item Require only easy-to-obtain real world data.
    \item Accurate ranking of policies based on their real world performance.
    \item Generally applicable to a wide range of robotic tasks. 
\end{enumerate}
To satisfy these requirements, we propose a simple-yet-practical approach that is composed of two components. The first component is out-of-distribution detection (OOD): we collect a set of real world observations that is independent of the trained policies, and evaluate whether this set is out-of-distribution with respect to the observations generated by DR and were used to train the policy. The idea is that if the DR robustness hypothesis indeed holds, then the real world should be seen as just another training environment, and therefore will not be identified as OOD. We find that a simple OOD method based on fitting a Gaussian mixture model (GMM) to the neural network activations already performs better than baselines based on offline policy evaluation~\cite{irpan2019offpolicy}. However, the performance can be sensitive to the particular choice of OOD parameters. Our second component is an evaluation in simulation of the learned policies on held out DR environment attributes, in the spirit of a validation set; ideally, a well trained DR policy will obtain high performance on this evaluation. Our method combines both components into a single score, and we therefore term it \textit{Validate on Sim, Detect on Real} (\methodname). Interestingly, we find that both components are complementary -- the combined score improves on each individual component, in an extensive study of diverse parameter settings. More importantly, we show that \methodname\ can provide accurate and stable policy rankings.

To evaluate our method, we perform an extensive study on a robotic manipulation task with image inputs. We train over 60 policies, using various RL and DR configurations, and collect over 130 hours of real-world evaluations. We then show that for various OOD methods and performance evaluation measures, \methodname\ consistently outperforms various baselines. We further evaluate the effect of policy variability on ranking accuracy, and compare several protocols for collecting validation sets of real-world observations and their effect on policy ranking. 

\section{Related work} 

Recent studies of offline RL proposed metrics to evaluate a policy based on data collected from running a different policy. Most of these studies do not consider the sim-to-real gap, and assume that only the policy differs between train and test~\cite{dudik2011doubly, jiang2016doubly, 10.5555/645529.658134, Thomas2015HighConfidenceOE, paine2020hyperparameter}. The closest work to ours is Off-Policy Classification (OPC \cite{irpan2019offpolicy}), which builds on actor-critic RL methods, and measures how accurately the trained critic classifies between successful or unsuccessful real trajectories. This method requires large amounts of very specific real-world data (successful vs. unsuccessful runs). In our experiments, even when such data was given, we found that our method significantly outperformed OPC.

Out-of-distribution detection, a.k.a. novelty detection or outlier detection, has been investigated extensively. Classical methods include one-class SVMs \cite{Scholkopf2001}
and support vector data description \cite{Tax:2004:SVD:960091.960109}, and more recent methods build on deep generative models~\cite{ZhaiCLZ16,schlegl2017unsupervised}. Recently, for image domains, several methods proposed to first train a deep network on some image classification task, and then learn a probability distribution of the feature activation in the network, as displayed in the training data. In \cite{lee2018simple}, a Gaussian distribution was used, while \cite{zisselman2020deep} extended this approach to flow-based distribution models. Here, we follow a similar approach, and learn the distribution of feature activations of the trained policy \textit{when run on training environments}. We focus on GMMs, which we found to work well, but our approach can be used with other models, such as the ones in \cite{lee2018simple, zisselman2020deep}.

\section{Background}
\subsection{Reinforcement learning} \label{background:rl}
We follow the Markov Decision Process (MDP) formulation of RL~\cite{sutton2018reinforcement}. Consider an agent that at every time step $t$ is in state $s_t \in  \mathcal{S}$, executes an action $a_t$ out of a set of possible actions $\mathcal{A}$, and transitions to a new state based on the transition probability $s_{t+1} \sim p(s_{t+1}|s_t, a_t)$. We denote the initial state distribution by $P_{init}(s)$. The agent also receives a reward $r_t = R(s_t, a_t)$. The goal is to find a policy $a_t = \pi(s_t)$ that maximizes expected cumulative reward, $\mathbb{E}\left[\sum^T_{t=1} \gamma^t r_t \right]$, where $\gamma\in(0,1]$ is the discount factor.
The value of a state-action pair when following a policy $\pi$, $Q^{\pi}(s,a)$,
is defined as the expected sum of discounted rewards
from that state-action pair:
$$
Q^{\pi}(s,a) = \mathop{\mathbb{E}}_{a \sim \pi} \left[ \sum^{\infty}_{k=t+1} \gamma^{k-t-1}r_k \Big\vert s_t=s, a_t=a \right].
$$ 
Actor-critic algorithms aim to find the optimal policy $\pi^*(s_t)$ by using two functions approximators, one for the policy $\pi_{\theta}(s_t)$ (actor) and the other for state-action value function $Q^{\pi}_{\phi}(s_t,a_t)$ (critic). Specifically, we use \textit{Soft Actor-Critic} (SAC) ~\cite{haarnoja2018soft, laskin2020reinforcement} for our robot learning experiments.

\subsection{Sim2Real transfer} \label{background:s2r}
Domain Randomization~\cite{tobin2017domain,james2019simtoreal,matas2018simtoreal,sadeghi2017sim2real} is a method for training neural network decision policies in simulation, such that their performance will transfer well to the real world. The idea is to apply random variations of the environment during the RL training. If these variations resemble the difference between simulation and reality, and the trained policy learns to be invariant to them, we expect that the policy will transfer well. 
More formally, we let $\phi$ denote the parameters of an environment. The parameters encapsulate all the factors of variation between domains. Let $P_{DR}(\phi)$ denote the distribution of domain parameters under domain randomization, and let $\phi_{real}$ denote the parameters of the real world. The hypothesis underlying domain randomization is that $P_{DR}(\phi_{real}) > 0$, that is, the real world is in the support of the domain randomization distribution.\footnote{This definition assumes discrete distributions. For continuous distributions, a similar condition is: there is some area $\phi_{real}$ where the performance of any policy does not change much, and $\int_{\phi_{real}} P_{DR}(\phi)d\phi > 0$.}

Various DR variations have been explored in the literature, such as manipulating textures, lighting, camera position, and physical properties. In general, however, which DR setting works best for a given domain is not known in advance, and an iterative design process is followed, where the DR variation is tweaked based on the real world performance.

Following \cite{dai2019analysing}, during training, a new domain is randomized every one or more simulation steps. We term this the \textit{frequency} of DR, and it is another hyper-parameter that can be tweaked.

\subsection{Out of distribution detection} \label{background:ood}
In this work, we build on the OOD approach of \cite{lee2018simple}, which exploits the data used for training a network on some task, to detect whether a test sample is in distribution or not. Given $M$ observations from the training domains $o_1,\dots,o_{M}$, we use them as input to the trained policy network, and collect the corresponding feature activations $x_1,\dots,x_{M}$. We then fit a parametric model $P_\theta(x)$ to the features of the training data $x_1,\dots,x_{M}$, using maximum likelihood. Finally, we use the parametric model as a score of how much a test input $o_{real}$ fits the observed data, by first extracting its features $x_{real}$, and then calculating the score $l = \log P_\theta(x_{real})$. Intuitively, for test inputs that are very different from the data, we expect the model to give a low likelihood, and therefore a low score.

Choosing which features of the network to include in the model, and which parametric model to fit, largely depends on the task and the data. \cite{lee2018simple} considered convolutional neural networks (CNNs), and in each layer, summed the activations that correspond to a particular channel, resulting in an activation that has the dimensionality of the number of channels in the layer. They fit a different Gaussian model for each layer, and used a weighted average of the models for the final score, where the weights were fit using labeled OOD examples. 

\section{Method} 
\label{s:method}

We now present our method for ranking policies based on combining validation in simulation and OOD detection in the real world. The method is summarized in Appendix~\ref{app:method_summary}.

\subsection{Validation in simulation} \label{ssec:EvalSim}
Following the DR hypothesis (that the real world can be seen as just another domain), a sufficient condition for a policy trained in simulation to perform well in the real world is that it performs well on every simulated domain within the support of $P_{DR}$. This can be measured by evaluating the policy on domains that were held out during training.

Formally, the training dataset for policy $\pi_i$ contains $N_{train}$ training domains $\phi_1,\dots,\phi_{N_{train}}$, sampled i.i.d.~from $P_{DR_i}$. We also collect another set of $N_{val}$ validation domains, sampled i.i.d.~from $P_{DR_v}$, which is used to evaluate all policies.
During training, the DR frequency (in simulation steps) in which a new domain $\phi_i$ is drawn is a hyper-parameter. However, in validation we only draw a new domain $\phi_k$ at the start of an episode and fix it for the remainder of the episode, similarly to the static real world setting.

Each policy is run in simulation under the above terms, and its \textit{validation score} (denoted by $s$) is its performance on the task, which can be measured by any suitable metric (the metrics we used are detailed in Section~\ref{ssec:TASK}). We obtain our \textit{prior ranking} by ranking all policies according to their validation scores.

\subsection{OOD detection in real world} \label{ssec:GMM}
The second component in \methodname\ evaluates how likely are observations from the real-world, with respect to the observations seen during training. Thus, this part builds on two techniques: (1) real world data collection, and (2) OOD detection. We first explain (2), and then discuss (1) in detail.

As mentioned earlier, we are inspired by \cite{lee2018simple}, and follow a similar OOD approach. We run each of the policies in the exact same setting it was trained on, and save a dataset of activations for each layer in the policy network. We then fit a $n$-component GMM ($n$ is a hyper-parameter) to each of the datasets (separate GMM per layer). In \cite{lee2018simple}, scores from several layers were combined using a weighted average, tuned on labeled OOD examples. However, in our work, we do not have such labeled outliers (we don't know in advance which real-world examples will be OOD for each policy). Therefore, we consider only single layers, and evaluate the GMM of each one separately (that is -- the choice of layer is a hyper-parameter). In our experiments, we verify that our results hold for different choices of layers.

We next discuss how to obtain real-world observations for evaluating the OOD score. Ideally, the observations should correspond to states that are relevant to the task, and are also visited by the policy in simulation (so that we measure OOD in the visual domain only). In our work, we consider 3 data collection protocols:
\begin{enumerate}
    \item \texttt{Expert}: Full demonstrations of an expert solving the task.
    \item \texttt{Sparse-expert}: A sparse set of observations sub-sampled from \texttt{Expert}. 
    \item \texttt{Initial}: Only initial episode observations, randomly sampled from $P_{init}$.
\end{enumerate}
All of the protocols are independent of the RL training method, and can be collected once and used for any trained policy. Obviously, Protocol (3) is easiest to generate, but for many tasks, teleoperation or scripted policies can be used to collect data for (1) and (2). We expect (1) to cover states that are relevant for the task, while not necessarily visited by the policy during training, and all the states in (3) to be visited by the policy, while possibly not covering all the important states in a successful trajectory. Protocol (2) is used to evaluate whether a complete trajectory is essential to our method or not. We note that for certain tasks, special care needs to be taken when generating data for protocol (1). For example, if an obstacle can be avoided in multiple ways, the data collected would need to include all of them in order to mitigate distribution shift between the data and the trained policies. Also, certain task setups may preclude use of protocol (3). For example, if the camera is mounted on the robotic arm, the initial observation may not be informative enough. In our experiments we used a fixed camera (see Section~\ref{ssec:SETUP}).

The \textit{OOD score} for a policy (denoted by $r$) is obtained by calculating the average GMM log-likelihood of all the observations in the real-world dataset.

\subsection{Merging the validation and OOD scores}\label{ssec:CMB}
So far we described how to produce two scores for a policy -- based on validation on sim, and OOD detection on real. We next describe how to combine the two scores into a single score.

The min-max normalization of a vector $V$ is defined as:
$$
\forall v \in V \quad v\text{’}=\frac{ v-\min{V}}{\max{V}-\min{V}}
$$

We denote the vectors of validation and OOD scores for all policies by $S$ and $R$, respectively. For a policy $i$, the \methodname\ score $VSDR_i$ is obtained by min-max normalization and multiplication of the individual scores:
$$
VSDR_i=s_i\text{’}\cdot r_i\text{’} \quad (s_i \in S, r_i \in R)
$$

The motivation for multiplying the normalized scores is that we want a good policy to excel \textit{in both} measures. 

\section{Results}
\label{RESULTS}
In this section, we evaluate our model selection approach.
We focus on answering the following questions:
\begin{enumerate}
    \item Does the combined \methodname\ score improve upon using either score (validation on sim and OOD detection on real) separately? 
    \item How does \methodname\ compare to baselines?
    \item How sensitive is our approach to different hyper-parameters selection?
\end{enumerate}

Obviously, items (1) and (2) depend on the task, the RL algorithm, the number of trained policies, the DR parameters, and the OOD method. To draw general conclusions, we opted to focus on a single, and rather generic, vision-based robotic grasping task, but performed an extensive evaluation of the different RL and DR parameters, trained policies, and OOD method. 
While our results are expected to vary significantly between the different parameter settings, we shall see that general observations about the method can still be drawn.

\subsection{Evaluated task - grasping}
\label{ssec:TASK}
We evaluate our method on a grasping task, based on the \texttt{Lift} task in Robosuite \cite{zhu2020robosuite}. The task goal is to grasp a 4 cm wide cube located on a table in front of the robot, lift it to a height of at least 4 cm and hold it at that height. Each grasping attempt is limited to $N_{steps}$ steps. For each grasping attempt, we calculate three performance metrics: \texttt{Reward}, \texttt{Success} and \texttt{Strict-Success}. \texttt{Reward} is the cumulative reward, taking into account both the reaching and grasping stages of the task. An attempt is deemed a \texttt{Success} if on at least one step the goal criteria was met. An attempt is deemed a \texttt{Strict-Success} if the goal criteria was met in some step $n_{succ}$, and continues to be met for all steps $n$ where $n_{succ} < n <= N_{steps}$. In our experiments, in training and evaluation, in both simulation and real-world, we set $N_{steps}=20$.

\subsection{Experimental setup}
\label{ssec:SETUP}
We use a 7DoF Franka Panda robotic arm in both simulation and the real-world, with a 4D Cartesian position-based impedance controller.

\textbf{Input:} The camera observations are retrieved from a fixed RGB camera (Intel RealSense D415) positioned in front of the robot looking  downwards at the table. The image is downscaled to 267x267 resolution and is then center cropped during evaluation, or randomly cropped during training into a 224x224 image, as suggested in RAD \cite{laskin2020reinforcement}. An image of the real-world setup can be seen in Figure \ref{fig:method_diagram}. The observation also includes a proprioceptive component, comprised of the sines and cosines of the joint angles, 3D end-effector Cartesian position and orientation (4D quaternion), and gripper finger positions. 

\textbf{Output:} Actions consist of the requested changes in 3D Cartesian position and in rotation of the gripper along the vertical axis,  and a binary gripper open/close command.

\textbf{Network Architecture:} We used an SAC \cite{haarnoja2018soft} agent, with neural network architecture similar to the one described in \cite{yarats2019improving} to solve the task in simulation. Unlike \cite{yarats2019improving}, we used a ResNet18 \cite{he2015deep} encoder to process a 224x224 image. As in \cite{yarats2019improving}, the image encoder is followed by actor and critic networks, both of which are 3-layer MLPs with ReLU activations following each layer except the last.

\textbf{Simulation:} We use the Robosuite simulation framework \cite{zhu2020robosuite}. We first sample two sets of domain datasets to be used during training and validation in simulation. All our policies are trained on domains from the training domains dataset, but are then evaluated on the validation domains dataset in order to get the validation score $s$. 

\textbf{Real World:} We fixed the lab lighting by shading the windows and using only artificial light for a fair comparison between all our tested policies. To control the robot we use ROS \cite{Quigley09ros} with Franka Panda libraries from \cite{sidhik2020}. To facilitate automated testing, we implemented a heuristic pick-and-place policy for placing the cube at desired positions.

\textbf{Policy Categories:} We trained policies using 20 distinct DR configurations. For each DR configuration we trained 3 seeds, for a total of 60 policies overall. We differentiate between 2 main configurations: \texttt{DR-heavy} and \texttt{DR-mild}. With \texttt{DR-heavy}, when a new object texture or material property is randomized, it completely replaces the previous one. With \texttt{DR-mild}, randomization is taken as a small perturbation around the original object texture and material. We used textures generated programatically by the simulator and real-world textures sampled from the DTD dataset \cite{cimpoi2014dtd}. DR configurations also varied by the randomization level of lighting parameters and by the frequency of applying new domains (once every 1/2/5 simulation steps and also once per episode at reset). Some examples are shown in Figure~\ref{fig:observations_lab_setup}. More details are provided in Appendix \ref{app:model_categories}.

\begin{figure}
    \centering
    \includegraphics[width=1\linewidth]{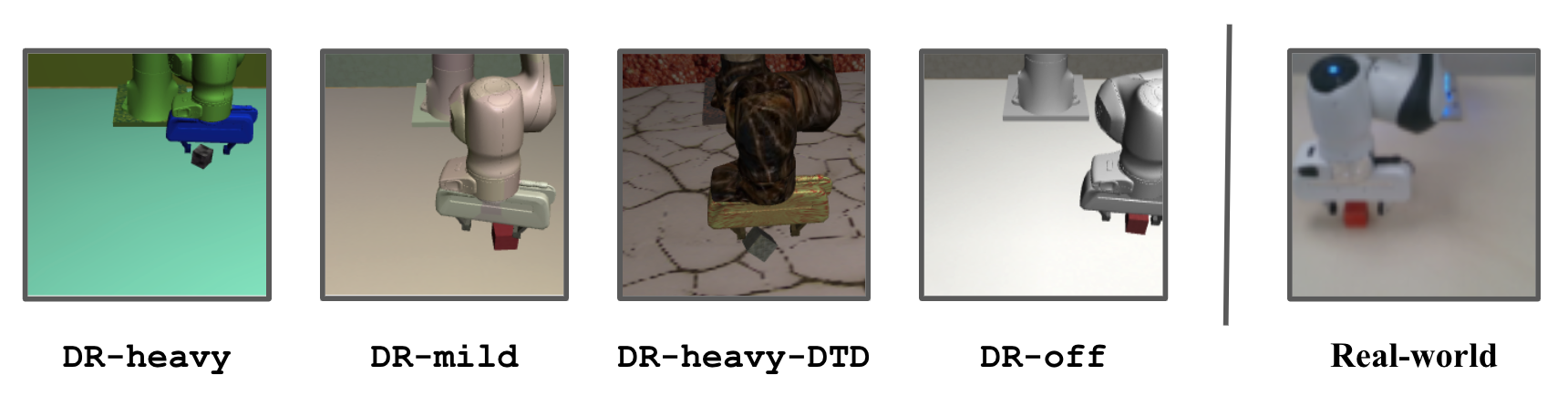}
    \caption{Examples of image observations. The 4 images on the left show different policy categories in simulation (see Appendix~\ref{app:model_categories}). The rightmost image shows an observation from our real-world experiments.}
    \label{fig:observations_lab_setup}
\end{figure}

\textbf{Baselines:} We compare \methodname\ to the SoftOPC and OPC metrics \cite{irpan2019offpolicy}.

\subsection{Experiments}
\label{ssec:EXP}

To measure the goodness of a model selection method, we compute the Spearman rank correlation coefficient $\rho$ \cite{spearman1904} between the ranking produced by the method and the ground-truth (GT) ranking. The GT ranking is the ranking of policies by their real-world performance. To evaluate real-world performance, each policy was tested on 49 grasp attempts (a 7x7 grid of initial cube positions). We generated 3 GT rankings, based on the \texttt{Reward}, \texttt{Success}, and \texttt{Strict-Success} metrics defined in Section~\ref{ssec:TASK}. Average real-world \texttt{Success} rate was 30\% across policies, ranging from 0\% to 92\%. To evaluate what factor of our results is attributed to noise in the real-world setting, we performed 2 independent evaluation cycles in the real-world (a total of approx. 130 robot hours) and calculated the Spearman $\rho$ between them -- this is an upper bound for any ranking method. Details are provided in Table~\ref{tbl:r2r_corr}.

\begin{table}
    \caption{Real-world run-to-run correlation, based on the metrics defined in Section~\ref{ssec:TASK}}
    \label{tbl:r2r_corr}
    \begin{center}
        \begin{tabular}{llll}
            \toprule
            Metric & Run 1 mean & Run 2 mean & Spearman $\rho$\\
            \midrule
            \texttt{Reward} & 15.49 & 15.39 & 0.98\\
            \texttt{Success} & 32.0\% & 30.1\% & 0.97\\
            \texttt{Strict-Success} & 17.6\% & 16.9\% & 0.93\\
            \bottomrule
        \end{tabular}
    \end{center}
\end{table}

We experiment with 3 types of validation in simulation, using the following DR configurations: \texttt{DR-heavy}, \texttt{DR-mild} and no DR (denoted by \texttt{DR-off}). For \texttt{DR-heavy} and \texttt{DR-mild}, the domains come from the validation domains dataset we collected in advance. A new domain is applied once per-episode at reset. Each sim validation score was computed based on 2,000 episodes.

For fitting the GMM, we consider features from five different layers in the policy network -- the output of the ResNet18 image encoder, referred to as \texttt{img\_encoder} and the first four layers in the actor MLP, referred to as \texttt{fc0}, \texttt{relu0}, \texttt{fc1} and \texttt{relu1}. We also experimented with GMMs with different number of components: 1, 2, 3, 5, 10.

For evaluating the real-world OOD scores, we collected a dataset of expert demonstrations, consisting of 64 grasp trajectories (grid of 8x8 initial cube positions), executed by a scripted policy. The total number of observations in this dataset is 704. We apply the data collection protocols defined in Section~\ref{ssec:GMM} to the expert dataset: For \texttt{Expert} we use the entire dataset. For \texttt{Sparse-expert} we uniformly sample 100 observations from the dataset. For \texttt{Initial} we take the initial step from each trajectory in the dataset, for a total of 64 observations.

To calculate the OPC and SoftOPC baselines, full trajectories are required, both successful and failing. \cite{irpan2019offpolicy} used 4000 real-world trajectories, generated by two policies with a success rate of approx. 40\%. To be able to calculate OPC and SoftOPC with a dataset of size of similar order, we used all the real-world trajectories we collected during GT evaluation of all trained policies, for a total of 2940 episodes (approx. 53k transitions), with a success rate of approx. 30\%. Note that this gives a very strong advantage to the baseline, which effectively trains on ground truth data!  Nevertheless, our results show that \methodname, which only has access to real-world data that does not depend on the trained policies, is able to outperform OPC and SoftOPC on our evaluated task.

\subsubsection{Main results and analysis}
\label{ssec:ANALYSIS}

\begin{figure}[t]
    \makebox[\textwidth][c]{\includegraphics[width=1.2\textwidth]{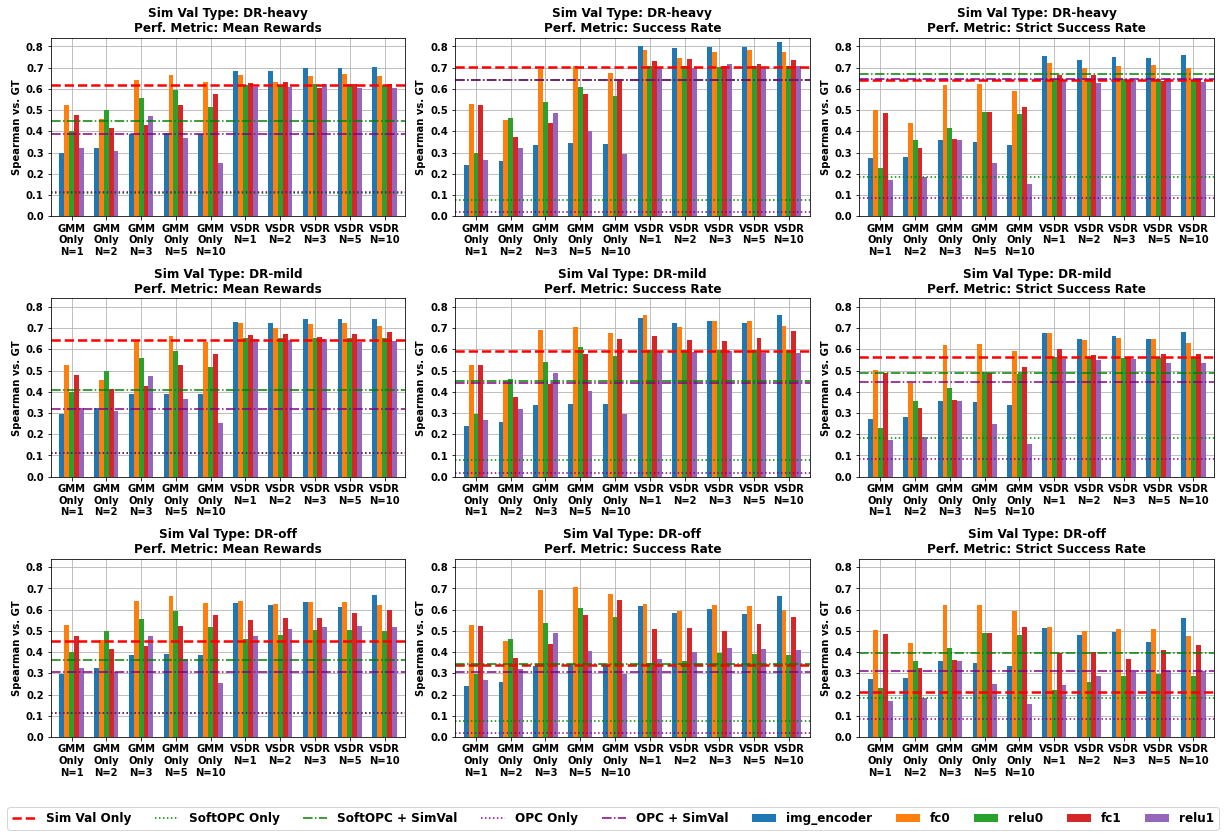}}
    \caption{Spearman correlations with GT rankings, for rankings based on different model selection methods. Each plot refers to a specific combination of simulation validation type and performance metric, indicated by the plot title. Note that each performance metric has its own GT ranking. Within each plot, the 5 bar groups on the left represent rankings based only on OOD detection in the real-world (i.e. GMM-only), and the 5 bar groups on the right represent VSDR-based rankings. In each bar group, ``N'' is the number of components used for the GMM. Each single bar refers to a different layer embedding used to fit the GMM. Horizontal lines represent correlations for rankings calculated by baseline methods (the "Sim Val Only" line in each plot corresponds to that plot's sim validation type). Note that within each column of plots, the GMM-only results are the same, yet we repeat them in each plot to allow easy comparison with VSDR results. }
    \label{fig:gmm_n_components_barplot}
\end{figure}

In Figure~\ref{fig:gmm_n_components_barplot}, we present the Spearman correlations vs. GT rankings achieved by \methodname\ and the baselines, for all combinations of $n$-component GMMs, chosen layer embedding for fitting the GMM, sim validation types and performance metrics: 225 different rankings overall.

We now turn to answer the questions we presented at the beginning of Section~\ref{RESULTS}.

First, we observe that \methodname\ does indeed improve upon the separate component scores in most cases. Compared to ranking according to validation in sim, \methodname\ rankings achieved better correlation vs. GT in 87\% of cases overall. Compared to ranking according to OOD detection on real (i.e. GMM-only rankings), \methodname\ rankings achieved better correlation in 86\% of cases overall. A more fine-grained analysis is given in Section~\ref{ssec:HYPER_PARAMS}.

Second, we evaluate how \methodname\ compares to baselines. Comparing to OPC and SoftOPC, \methodname\ based rankings achieved better correlation in 100\% of cases, despite the fact that VSDR is shown far less data. As a side note, we see that combining the sim validation score with OPC/SoftOPC improves each of these metrics as well (though it does not improve upon sim validation rankings in most cases). This shows that the idea of combining validation in sim with performance in real generalizes beyond our OOD method. Even so, \methodname\ still produces the best correlation in most cases (92\% compared to sim val + OPC and 88\% compared to sim val + SoftOPC).  It is important to note, in any case, that in these results OPC/SoftOPC still use GT data, and therefore these combined methods are not practical for our model selection desiderata as presented in Section~\ref{s:intro}.

\subsubsection{Analysis of hyper-parameter choices}
\label{ssec:HYPER_PARAMS}

We now evaluate the sensitivity of our approach to different hyper-parameter selections.

\textbf{Data collection protocol:} The results in Figure~\ref{fig:gmm_n_components_barplot} were obtained using the \texttt{Sparse-expert} data collection protocol, that is~- with 100 real-world observations. Results using \texttt{Expert} (704 observations) and \texttt{Initial} (64 observations) were similar (see Appendix \ref{app:data_col_protocols} for details). This highlights the small amount of data sufficient for our method, and also shows that complete trajectories are not a requirement. In addition, we observed that sub-sampling different sets of observations as part of the \texttt{Sparse-expert} protocol did not have a significant impact on the results. In our experiments, the standard deviation for the correlations based on the \methodname\ score obtained from different sub-sampled sets, ranged between 0.0004 to 0.02. Collectively, these results display the robustness of our approach to different real-world data collection approaches.

\textbf{Choice of NN layer for fitting the GMM:} Here we make a couple of observations: (1) Choosing to fit the GMM to ReLU feature activations seems to generate VSDR rankings which are slightly worse than the ones generated using FC layers. More concretely, out of the 30 cases in which the VSDR-ranking had worse correlation compared to the sim val ranking, 26 were with a GMM fit to one of the ReLU layers' activations ; (2) Fitting the GMM using feature activations from earlier layers in the NN (namely \texttt{img\_encoder} and \texttt{fc0} in our case) tends to lead to better rankings compared to feature activations from layers closer to the NN output. Comparing to sim val rankings again, VSDR rankings based on GMMs fitted to either \texttt{img\_encoder} or \texttt{fc0} improved the Spearman correlation by an average of 38\%. With GMMs fitted to subsequent layers, the improvement was only 11\% on average.

\textbf{Validation in simulation settings:} Validating in simulation with richer, more complex randomization (\texttt{DR-heavy}) tends to lead to a better VSDR ranking. Harder and more diverse domains are better at separating the policies from one another, and are a better predictor of crossing the sim2real gap. This is also noticeable when comparing to the separate component scores. Looking at Figure~\ref{fig:gmm_n_components_barplot}, a couple of observations can be made: (1) Comparing VSDR to sim val only rankings, it is clear that the harder the DR setting, the less it benefits from combining with the OOD (GMM-only) score. Concretely, with \texttt{DR-off}, the improvement from VSDR over sim val only is 53\% on average with a maximum of 162\%. With \texttt{DR-mild} the average is 8\% and maximum is 29\%. With \texttt{DR-heavy} the average is 5\% and maximum is 19\%. (2) Comparing to GMM-only rankings, VSDR rankings based on \texttt{DR-heavy} and \texttt{DR-mild} are almost always better (99\% of cases). On the other hand, when using \texttt{DR-off}, VSDR rankings often fail to improve over the GMM-only rankings (improvement in only 60\% of cases).

\textbf{Number of GMM components}: From Figure~\ref{fig:gmm_n_components_barplot}, we can observe that the number of components used has a significant effect when ranking policies based on GMM fitting only, with 1- and 2-component GMMs performing noticeably worse. However, when combined with validation in simulation scores to generate VSDR scores, the number of components in the GMM has a much smaller effect. 

Overall, our empirical results indicate that the combination of sim validation and OOD scores is not simply additive, but complimentary - as intended. The information held in sim validation performance is evidently strong enough, that when combined with GMM scores, it overshadows minor trends that are visible in GMM-only rankings. On the other hand, we've shown earlier that VSDR rankings outperform sim validation rankings in 87\% of cases, and especially when the sim validation rankings are weak (as is the case with \texttt{DR-off}). All of which goes to indicate that OOD detection in the real-world provides useful information on top of sim validation performance, as well.

Additional results are shown in Appendix~\ref{app:additional_results}.

\section{Conclusions}
We proposed a method for choosing the best policies to deploy in the real-world that are likely to best cross the sim-to-real gap. Our method requires merely dozens of real-world observations in order to evaluate how novel is the real world domain comparing to the domains each policy has observed during simulation, and outperforms baselines from the literature. 
In this work we focused on image observations, and DR methods that modify the visual appearance of the scene. It is interesting to extend this approach to domains where physical properties such as mass and friction are varied using DR as well.

\section*{Acknowledgments}
Aviv Tamar is partly funded by the the Open Philanthropy Project Fund --
an advised fund of Silicon Valley Community Foundation, and a grant from Intel Corporation.

\small{
  \bibliography{main}
  \bibliographystyle{abbrvnat}
}

\appendix

\section{Appendix}

\subsection{Method summary}
\label{app:method_summary}

Our method, described in detail in Section~\ref{s:method}, is summarized in Algorithm \ref{alg1}.

\begin{algorithm}[H]
\caption{Validate on Sim, Detect on Real Score}\label{alg1}
\begin{algorithmic}[1]
\Require Policies $\pi_1,\dots,\pi_k$, trained using different DR parameters $P_{DR_1}(\phi),\dots,P_{DR_k}(\phi)$, single set of DR parameters for validation $P_{DR_v}(\phi)$, real world observations dataset $\mathcal{D}_{R}$, number of GMM components $N$, layer in policy network $L$

\For{$i=1,\dots,k$} 
    \State Run $\pi_i$ in sim $P_{DR_v}(\phi)$ to get prior sim validation score $s_i$
    \State Run $\pi_i$ in sim $P_{DR_i}(\phi)$ to get feature activations $\mathcal{A}_{S}$ from layer $L$
    \State Fit a $N$-component GMM $G_{i}$ to $\mathcal{A}_{S}$
    \State Feed-forward $\mathcal{D}_{R}$ on $\pi_i$ to get feature activations $\mathcal{A}_{R}$ from layer $L$
    \State Calculate log-likelihood of $\mathcal{A}_{R}$ on $G_{i}$ to get OOD score $r_i$ 
\EndFor
\For{$i=1,\dots,k$} 
    \State Min-max normalize $s_i\text{’}, r_i\text{’}\leftarrow  s_i, r_i$ 
    \State $VSDR_i\leftarrow s_i\text{’}\cdot r_i\text{’}$
\EndFor
\State\Return $VSDR_1,\dots,VSDR_k$ 

\end{algorithmic}
\end{algorithm}

\subsection{Policy categories}
\label{app:model_categories}

We trained policies using a variety of domain randomization configurations, which we divide into the following categories:
\begin{itemize}
  \item \texttt{DR-heavy-freq-n:} For each object in the scene: Randomize the textures (programatically generated by the simulator) and material properties. For each lighting source: Randomize activation, position, direction and diffuse, ambient and specular properties. For each camera: Randomize position, angle and field-of-view. \texttt{freq-n} refers to the simulation steps frequency in which a new domain is applied, where $n \in \{0,1,2,5\}$. $n=0$ stands for randomizing a new domain once per episode at reset.
  \item \texttt{DR-mild-freq-n:} Same as \texttt{DR-heavy-freq-n}, but the randomization is taken as a small perturbation around the original object texture and material.
  \item \texttt{DR-dtd:} Same as \texttt{DR-heavy-freq-1}, with object textures sampled \textbf{only} from the DTD Dataset \cite{cimpoi2014dtd} (no programatically generated textures).
  \item \texttt{DR-heavy-dtd:} Same as \texttt{DR-heavy-freq-1}, with both programatically generated textures and textures from DTD. The texture has a 20\% chance of being sampled from DTD.
  \item \texttt{DR-heavy-lights-always-on:} Same as \texttt{DR-heavy-freq-1}, only that the lights are never turned off. 
  \item \texttt{DR-heavy-light-params-sweep:} Same as \texttt{DR-heavy-freq-1}, with stronger randomization of the diffuse, ambient and specular properties of each light source. We increase the randomization for each property separately and for all combinations, for a total of 8 policies.
  \item \texttt{DR-off:} No randomization is applied. 
\end{itemize}
Examples of image observations generated for some of the configurations are shown in Figure \ref{fig:observations_lab_setup}.

\subsection{Additional results}
\label{app:additional_results}

\subsubsection{Comparing data collection protocols}
\label{app:data_col_protocols}
In Section~\ref{ssec:GMM} we define 3 data collection protocols for obtaining the real-world observations dataset on which we evaluate the OOD score: \texttt{Expert}, \texttt{Sparse-expert} and \texttt{Initial}. The results shown in Figure~\ref{fig:gmm_n_components_barplot} were obtained using \texttt{Sparse-expert}. We show that, on aggregate, the results obtained using all data collection protocols are similar. For each data collection protocol, we calculate the \methodname\ based ranking for all combinations of simulation validation types, performance metrics, and layer for fitting the GMM (for the results in this section we used a 2-component GMM). For each \methodname\ based ranking we calculate the Spearman correlation with the corresponding GT ranking. In Figure~\ref{fig:compare_ood_data_collections} we show box plots generated from all correlations per data collection protocol.

\subsubsection{Policies separability}
\label{app:models_separability}
\begin{figure}[]
  \centering
  \makebox[\textwidth][c]{\makebox[1.2\textwidth]{
  \begin{minipage}[]{0.55\textwidth}
    \includegraphics[scale=0.4]{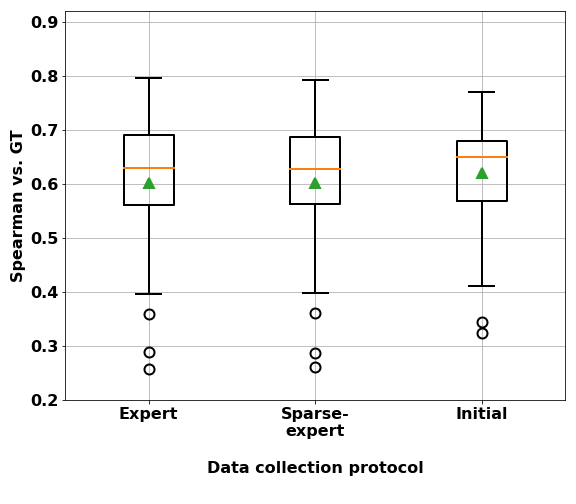}
    \captionsetup{justification=justified}
    \caption{ Box plots of Spearman correlations of \methodname\ based rankings with GT rankings, for each of the data collection protocols defined in Section~\ref{ssec:GMM}. }
    \label{fig:compare_ood_data_collections}
  \end{minipage}
  \hfill
  \begin{minipage}[]{0.55\textwidth}
    \includegraphics[scale=0.4]{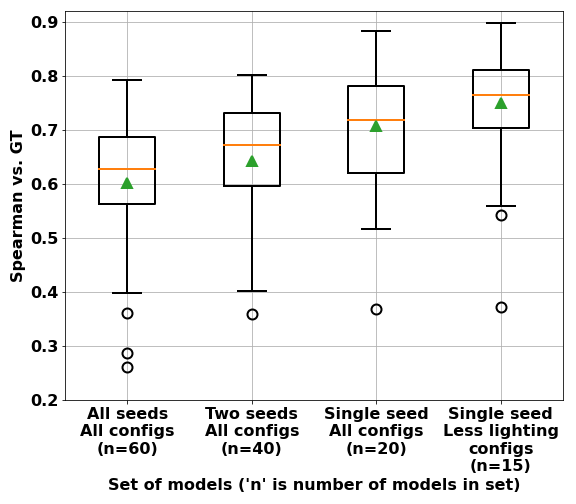}
    \caption{Box plots of Spearman correlations of \methodname\ based rankings with GT rankings, across different sets of policies. Each plot going in the right direction is based on a smaller, more separable set of policies. }
    \label{fig:num_models_box_plot}
  \end{minipage}}}
\end{figure}

We show how correlation with the GT improves when the set of policies is more separable. Two approaches for making the policies more separable are: (1) Having fewer seeds per policy type; (2) Removing very similar policy types. In Figure~\ref{fig:num_models_box_plot} we show results for both approaches -- first we remove seeds, and then we remove 5 out of the 8 \texttt{DR-heavy-light-params-sweep} policies, keeping only the DR types where a single lighting property was changed. Box plots generated in a similar fashion to what is described in Appendix~\ref{app:data_col_protocols}.

\end{document}